\documentclass[
]{ceurart}

\usepackage{listings}

\usepackage{amsmath}
\usepackage{amssymb,amsfonts,bm}

\usepackage{makecell}

\begin{document}

\copyrightyear{2022}
\copyrightclause{Copyright for this paper by its authors.
  Use permitted under Creative Commons License Attribution 4.0
  International (CC BY 4.0).}

\conference{Forum for Information Retrieval Evaluation, December 9-13, 2022, India}

\title{Overview of the HASOC Subtrack at FIRE 2022: Offensive Language Identification in Marathi}


\author[1]{Tharindu Ranasinghe}[%
email=tharindu.ranasinghe@wlv.ac.uk,
]
\cormark[1]
\address[1]{University of Wolverhampton, United Kingdom}
\address[2]{George Mason University, United States}

\author[2]{Kai North}[%
]

\author[1]{Damith Premasiri}[%
]

\author[2]{Marcos Zampieri}[%
]

\cortext[1]{Corresponding author.}

\begin{abstract}
The widespread of offensive content online has become a reason for great concern in recent years, motivating researchers to develop robust systems capable of identifying such content automatically. With the goal of carrying out a fair evaluation of these systems, several international competitions have been organized, providing the community with important benchmark data and evaluation methods for various languages. Organized since 2019, the HASOC (Hate Speech and Offensive Content Identification) shared task is one of these initiatives. In its fourth iteration, HASOC 2022 included three subtracks for English, Hindi, and Marathi. In this paper, we report the results of the HASOC 2022 Marathi subtrack which provided participants with a dataset containing data from Twitter manually annotated using the popular OLID taxonomy. The Marathi track featured three additional subtracks, each corresponding to one level of the taxonomy: Task A - offensive content identification (offensive vs. non-offensive); Task B - categorization of offensive types (targeted vs. untargeted), and Task C - offensive target identification (individual vs. group vs. others). Overall, 59 runs were submitted by 10 teams. The best systems obtained an F1 of 0.9745 for Subtrack 3A, an F1 of 0.9207 for Subtrack 3B, and F1 of 0.9607 for Subtrack 3C. The best performing algorithms were a mixture of traditional and deep learning approaches.
\end{abstract}

\begin{keywords}
  offensive language \sep
  hate speech \sep
  text categorization \sep
  Marathi
\end{keywords}

\maketitle

\section{Introduction}


Social media is the main source of information, current events, and public opinion for many individuals \cite{Karlsen_Aalberg_2021}. Users typically spend on average 2 hours per day on social media platforms, such as Twitter, Facebook, and Instagram \cite{Riehm_etal_2019}. The impact social media has on an individual's perception of the world, as well as their mental well-being, is substantial \cite{bannink2014cyber}. Social media often contains hateful and offensive content, including but not limited to pejorative language \cite{dinu-etal-2021-computational-exploration}, cyber-bullying \cite{rosa2019automatic, yao2019cyberbullying, Shetty_etal_2021}, online extremism \cite{Saja_etal_2021}, and racism \cite{Kwok2013, tulkens2016dictionary}. Such content has been shown to cause anxiety or depression \cite{bannink2014cyber}, incite self-harm \cite{Johnetal2018}, or violence towards others \cite{Matthewetal2019}. 


With governments realizing the impact social media has on public health and opinion, social media companies have invested in ways to cope with hateful and offensive content \cite{Matthewetal2019, Akins2021}. To achieve this, they are interested in the development of machine learning (ML) models to automatically detect such content on their social media platforms. These models have traditionally consisted of Naive Bayes (NB) \cite{Dinakar2011, Chen_etal_2012, Kwok2013}, support vector machines (SVMs) \cite{Salminen_etal_2018, Jun-Ming_etal_2012, Dadvar_etal_2013, Nobata_etal_2016}, and random forests (RFs) \cite{Salminen_etal_2018, Nugrohoetal2019} while, more recently, neural networks \cite{Wang2019YNU, Bashar_Nayak2020, Saha_etal_2019,hettiarachchi-ranasinghe-2019-emoji}, and transformer-based models \cite{Saha_etal_2019, Kumar2020,ranasinghe2019brums, sarkar2021} have been applied to this problem. 

As evidenced by recent international competitions on this topic \cite{kumar2018benchmarking, offenseval, mandl2019overview, basile2019semeval, mandl2020, zampieri-etal-2020-semeval, kumar-etal-2020-evaluating}, a known challenge in detecting offensive and hateful content is the intrinsic subjectivity of such phenomena. In highly multilingual countries, such as India, this task becomes even more challenging. India has 22 official languages, many of which do not have large annotated datasets that can be used for the training and evaluation of hate speech and offensive language identification models. Marathi is an example of such an under-resourced language, especially when in comparison to high-resourced languages, such as Hindi or English \cite{gaikwad2021cross}.


This paper presents the results of Subtrack 3 at HASOC-2022 \cite{hasoc2022mergeoverview}: Offensive Language Identification in Marathi. HASOC-2022 challenged participating teams with the development of state-of-the-art hate speech and offensive language identification models. Subtrack 3 at HASOC-2022 aimed to encourage the development of such models for Marathi and thus encourage more research into the future development of models for under-resourced languages. 

This paper is structured as follows. Section \ref{related_work} introduces previous hate speech and offensive language shared-tasks. Section \ref{task} introduces Subtrack 3. Section \ref{data} describes the data used for this subtrack. Section \ref{approaches} lists the approaches adopted by each participating team. Section \ref{results} provides each teams' performance. Section \ref{conclusion} gives a conclusion on what is the current state-of-the-art for identifying hate speech and offensive language in Marathi and makes predictions regarding the direction of future research.

\section{Related Work} \label{related_work}


Numerous shared-tasks have been organized with the goal of providing benchmarks for the automatic identification of hateful and offensive content. These shared-tasks have consisted of differing forms classification, being either binary, multi-class, or multi-label classification, and contained differing subtracks aimed at predicting the type or target of the offensive content as well as dealing with offensive content in languages other than English. 

\subsection{Related Tasks}




\paragraph{\textbf{TRAC 2018-2022}} The first Trolling, Aggression and Cyberbullying (TRAC-2018) workshop hosted at COLING \cite{kumar2018benchmarking}, challenged participating teams to create multi-class classifiers for predicting which of three levels of aggression is depicted by a Facebook comment: overt aggression, covert aggression, or nonaggression. Their dataset consisted of randomly sampled English and Hindi Facebook comments.  Several issues were reported to the organizers regarding dataset quality:  1). some instances contained code-mixing between English and Hindi, and 2). some instances were considered incorrectly labelled due to their subjectivity. The second TRAC workshop (TRAC-2020) hosted at LREC \cite{kumar-etal-2020-evaluating}, introduced a new language, Bengali, as well as a new domain, Youtube comments, to resolve the previous issues in data quality. TRAC-2020 also provided a new subtrack in the form of binary classification of gender: gendered (GEN) or non-gendered (NGEN) instances. The third TRAC workshop (TRAC-2022) \cite{trac-2022-threat} expanded this subtrack further to include the recognition of other types of offensive content, including racial bias, religious intolerance, and casteist bias, as well as the identification of the discursive role of a given instance in relation to a previous comment. It also introduced a new domain, “COVID-19 related conversation” \cite{trac-2022-threat}. Performances on TRAC-2018 varied with neural networks performing on par with other approaches, such as SVMs and RFs \cite{kumar2018benchmarking}. TRAC-2020 and 2022, however, saw transformer-based models achieve the best performances \cite{kumar-etal-2020-evaluating}; teams that used an ensemble of transformers performed exceptionally well in comparison to other approaches.


\paragraph{\textbf{HatEval 2019}} Task 5 at SemEval-2019: Multilingual Detection of Hate Speech Against Immigrants and Women in Twitter (HatEval) \cite{basile2019semeval}, contained two subtracks . Subtrack A was a binary classification task. It saw  English or Spanish tweets be classified into hatespeech or non-hatespeech instances. Subtrack B was evaluated as a multi-label classification task which required the identified instances of hatespeech to be further grouped in regards to their “aggressive attitude and the target harassed” \cite{basile2019semeval}.  The organizers suggested that participating teams divide this task into two binary classifications tasks, whereby each instance's target is classified as being aggressive or non-aggressive, and its attitude is classified as being either a targeted message (individual) or a hateful message in general (generic). The first and second top performing models for Subtrack A were both SVMs having outperformed neural networks and other approaches. Performances on Subtrack B were noticeably worst than Subtrack A. The organizers contributed this to its multi-labelled nature.





\paragraph{\textbf{OffensEval 2019-2020}} Task 6 at SemEval-2019: Identifying and Categorizing Offensive Language in Social Media (OffensEval-2019) \cite{offenseval}, presented three subtracks to participating teams ranging for binary to multi-class classification on the Offensive Language Identification Dataset (OLID) consisting of English tweets. Subtrack A saw the binary classification of tweets with non-offensive and offensive labels. Subtrack B and C was concerned with identifying the target of the offensive tweet. Subtrack B was the binary classification of instances into either targeted or untargeted insults, whereas Subtrack C was the multi-class classification of instances into three labels concerned with the target type: “individual, group, or other” \cite{offenseval}. Task 12 at SemEval-2020: Multilingual Offensive Language Identification in Social Media (OffensEval-2020) \cite{zampieri-etal-2020-semeval} was an extension of the previous shared-task. It included the same subtracks as SemEval-2019, but presented additional tracks and data in Arabic, Danish, English, Greek, and Turkish. Both shared-tasks saw transformer-based models achieve the best results, such as BERT, RoBERTa, and mBERT. The multilingual nature of OffensEval-2020 also allowed for cross-lingual transfer learning, which was adopted by the highest performing systems.




\subsection{HASOC}

\paragraph{\textbf{HASOC 2019-2020}} The Hate Speech and Offensive Content
Identification in Indo-European Languages track (HASOC-2019) at FIRE \cite{mandl2019overview}, held three subtracks. Data was collected from Twitter and Facebook in Hindi, German, and English. Subtrack A was a binary classification task to distinguish between hateful and non-hateful instances. Subtracks B and C were more fine-grained classification tasks to identify the type (hate speech, offensive, and profanity) and target (targeted or untargeted) of each instance respectively. HASOC-2020 \cite{mandl2020} saw the removal of Subtrack C and the creation of a new dataset with the aim of minizing bias in the data collection process. This was achieved through the adoption of a more randomized sampling technique that minimizied the selection of types of hate-speech known to the dataset's creators \cite{mandl2020}. Neural networks achieved the best performances at HASOC-2019, with an LSTM model achieving the highest performance for English \cite{Wang2019YNU}. BERT-based models become more popular at HASOC-2020 and achieved the best performance for German. However, LSTM-based models again reported the best results for Hind and English. Overall performances were considered to be worst for HASOC-2020 than they were for HASOC-2019 \cite{mandl2020}. Organizers' reported this as being due to the more realistic sampling technique used in HASOC-2020.

\paragraph{Marathi at HASOC-2021} HASOC-2021 \cite{hasoc2021mergeoverview} included the same two subtracks as HASOC-2020 for English and Hindi, but was the first to apply these subtracks to Marathi. The data used for Marathi was extracted from the \textit{MOLD} dataset \cite{gaikwad2021cross}. The \textit{MOLD} dataset consists of 2,547 hateful and non-hateful tweets extracted from Twitter's API. Its hateful tweets were obtained by searching for keywords “related to politics, entertainment, and sports along with the hashtag \#Marathi” \cite{gaikwad2021cross}. All instances were then labelled by six Marathi speakers. The best performing models for Marathi used cross-lingual transfer learning \cite{nene-2021-transformer}. \citet{nene-2021-transformer} trained an XLM-R model on pre-existing data in Hindi from HASOC 2019. They then used the returned model's updated weights and softmax activation layer to initialize the weights of a new XLM-R model on the provided Marathi dataset. The second best performing model adopted a similar approach. \citet{Hasoc_2021_team91} fine-tuned the monolingual LaBSE and multilingual XLM-R models on Marathi and Hindi data and discovered than the multilingual XLM-R model achieved the best results. The other top-10 performing models submitted to HASOC-2021's Marathi subtrack included feature engineering approaches using word and character level n-grams \cite{bestgen_2021, HaCohen-Kerner_2021}, an LSTM model \cite{Velankar2021HateAO}, as well as other transformer-based models, such as mBERT and DistilmBERT \cite{Bhatia2021OneTR}. However, these models failed to achieve performances on par with those models fine-tuned on both Hindi and Marathi data.

\section{HASOC-2022} \label{task}
\subsection{Subtrack 3: Offensive Language Identification in Marathi}

HASOC-2022 hosted three subtracks with subtracks 1 and 2 dealing with binary and multi-class classification of German and the code-mixed language, Hinglish, respectively. Subtrack 3 focused on the automatic identification of offensive language in Marathi. Subtrack 3 was further divided into three additional subtracks. 

\begin{itemize}
    \item {\bf Subtrack 3A:} A binary classification task classifying posts into two classes: offensive (OFF), and non-offensive (NOT). 
    \item {\bf Subtrack 3B:} A binary classification task identifying whether a post is either targeted (TIN) or untargeted (UNT). 
    \item {\bf Subtrack 3C:} A more fine-grained classification task, whereby the previously identified TIN posts were further classified into individual (IND), group (GRP), or other (OTH). 
\end{itemize}

\section{Data} \label{data}

The data used for Subtrack 3 consisted of extracts taken from \textit{MOLD 2.0} \cite{Zampieri2022}. \textit{MOLD 2.0} is an expansion of the original MOLD dataset (v1.0) \cite{gaikwad2021cross}. It contains 3,600 tweets with an additional 1053 instances on top of the original instances provided by \textit{MOLD} (v1.0) \cite{gaikwad2021cross}. All instances were annotated using the popular OLID taxonomy \cite{OLID} with additional information being provided in regards to the target (level B) and target type (level C) of the offensive or non-offensive tweet. The annotation was carried out by three native speakers of Marathi with an inter-annotator agreement of 0.79 Cohen’s kappa across all instances. 

The \textit{MOLD 2.0} dataset was split into an 80/20 split between train and test sets, respectively. An unequal number of offensive (1067) and non-offensive (2034) instances were provided to the participating teams resulting in the label distribution shown in Table \ref{tab_labels}.


\begin{table}[!ht]
\centering
\scalebox{1.00}{
\begin{tabular}{cccrr|r}
\toprule
\bf A & \bf B &  \bf C & \bf Training & \bf Test & \bf Total \\ 
\midrule

OFF  &  TIN  &  IND &  503 & 51 & 554 \\
OFF  &  TIN  &  OTH &  80 & 56 & 136 \\
OFF  &  TIN  &  GRP &  157 & 51 & 208 \\ 
OFF  &  UNT  &  --- &  327 & 102 & 429 \\ 
NOT  &  ---  &  --- & 2,034  & 250 &  2,284 \\
\midrule
\bf All &    &     &  3,101  & 510 &  3,611 \\ 
\bottomrule 
\end{tabular}
}
\vspace{2mm}
\caption{MOLD v2.0 - Distribution of labels. Table taken from \cite{Zampieri2022}.}
\label{tab_labels}
\end{table}

\section{Approaches} \label{approaches}

In total, 8 out of the 10 participating teams submitted system reports for Subtrack 3. 10 teams took part in Subtrack 3A, 7 in Subtrack 3B, and 6 in Subtrack 3C. System descriptions are provided below.












\subsection{Traditional}

\paragraph{\textbf{ssncse\_nlp}} \citet{Hasoc_2022_ssncse_nlp}  used Scikit-learn's CountVectorizer function \cite{scikit-learn} to generate vectors of term/token counts. They fed these vectors as features into four types of traditional classifier: RF, SVM, logistic regressor, and a KNN classifier. They discovered that their RF model achieved the greatest performance and submitted this model for all three subtracks.

\paragraph{\textbf{satlab}} \citet{Hasoc_2022_satlab} submitted a simplified version of their previous model used for HASOC-2021 \cite{hasoc2021mergeoverview}. They trained a logistic regressor taken from the LIBLinear package \cite{LIBLINEAR} on several character-level n-gram  features. These features consisted of n-gram length, sentence position, and BM25 frequency. All n-grams with a frequency of less than two and over five characters in length were not considered. Feature scores were normalized per L2 regularization, and parameter optimization was conducted using four-fold cross validation over several random grid searches. This model was submitted to all three subtracks.

\paragraph{\textbf{ml\_ai\_iiitranchi}} \citet{Hasoc_2022_ml_ai_iiitranchi} experimented with a logisitic regressor, SVM, and RF. They lemmatized each instance and trained their models on token frequencies obtained using Scikit-learn's TF-IDF-Vectorizer function \cite{scikit-learn}. Their logistic regressor surpassed the performances of the other three models for Subtrack 3A, whereas their SVM achieved the best performance for subtracks 3B and 3C. They submitted their best performing models to each of the subtracks, respectively.

\subsection{Deep Learning}

\paragraph{\textbf{optimize\_prime}} \citet{Hasoc_2022_Optimize_Prime} compared the performances of various BERT-based models taken from HuggingFace that have been pre-trained on Marathi data  \cite{Zampieri2022}: MuRIL \cite{khanuja2021muril}, MahaTweetBERT \cite{patankar2022spread}, MahaTweetBERT-Hateful \cite{patankar2022spread}, and MahaBERT \cite{joshi-2022-l3cube}. They cleaned each instance provided by the shared-task of placeholder texts, such as “\textit{@USER}”, and removed hastags, URLs, empty parentheses and new lines. Each instance was then automatically encoded and then fed into each model. The AdamW optimizer was used during training with a learning rate of 1e-5 and batch size of 32. The training was conducted over 25 epochs. The MahaTweetBERT \cite{patankar2022spread} outperformed all of models when trained on a combination of instances from HASOC-2021 \cite{hasoc2021mergeoverview} and HASOC-2022 \cite{hasoc2022mergeoverview}. This model was only submitted for Subtrack 3A.

\paragraph{\textbf{hate-busters}} \citet{Hasoc_2022_hatebuster} treated Subtrack 3 as a multi-class classification problem instead of a multi-label or several binary classification tasks, and attempted to predict all possible labels at once (as shown in Table \ref{tab_labels}). They applied five and ten-fold cross validation by splitting the training and test sets into \textit{n} independent sets and trained an ensemble of \textit{n} separate models. The final output label was derived from plurality voting. They tested two different types of model: InfoXLM \cite{chi-etal-2021-infoxlm} and L3Cube \cite{joshi-2022-l3cube}. The former being trained on cross-lingual representations of 100 languages, whereas the later being trained solely on Hindi and Marathi data. The more language-specific L3Cube model achieved the greatest performance in their ensemble architecture. This ensemble was submitted for all three subtracks. 

\paragraph{\textbf{sakshi\_hasoc}} \citet{Hasoc_2022_sakshi} fine-tuned mBERT \cite{devlin2019bert}, DistilmBERT \cite{Sanh2019DistilBERTAD}, and MuRIL \cite{khanuja2021muril} on a resampled version of the provided training set to compensate for the imbalance between offensive and non-offensive instances. They oversampled the training set by including duplicates of offensive instances that made for a 0.5 ratio between the two labels. They achieved this using the RandomOverSampler function in the imblearn library \cite{imblearn}. MuRIL was found to achieve the best performance when trained over 4 epochs and set to a learning rate of 1e-4 and a batch size of 4. They submitted this model to Subtrack 3A.

\paragraph{\textbf{irlab@iitbhu}} \citet{Hasoc_2022_IRLab@IITBHU} removed usernames, punctuation, URLS, and emojis. They then concatenated the tweet, its comments, and responses and fed these concatenated instances into Fast.ai \cite{Howard_2020}, being an off-the-shelf deep learning package. They submitted their Fast.ai model to all three subtracks.

\paragraph{\textbf{citk\_isi}} \citet{Hasoc_2022_CITK_ISI} compared the performance of mBERT \cite{devlin2019bert} and MahaBERT \cite{joshi-2022-l3cube} with the latter being pre-trained on the L3 Cube-MahaCorpus. Each model was trained with a learning rate of 1e-5 and a batch size of 5 over 5 epochs. MahaBERT was found to outperform mBERT and was submitted for all three subtracks.

\section{Results} \label{results}


Team performances are displayed within tables \ref{tab:3A_results_table} to \ref{tab:3C_results_table} and are discussed throughout the following section.

\begin{table}[ht]
    \centering
    \scalebox{0.92}{
    \begin{tabular}{c|c|c|c|c|c}
        \hline
        \textbf{Rank} & \textbf{Name} & \textbf{Number of Runs} & \textbf{Precision} & \textbf{Recall} & \textbf{F1} \\
        \hline
        
        1 & ssncse\_nlp \cite{Hasoc_2022_ssncse_nlp} & 4 & 0.9758 & 0.9741 & 0.9745\\
        2 & optimize\_prime \cite{Hasoc_2022_Optimize_Prime} & 2 & 0.9588 & 0.9589 & 0.9588 \\
        3 & hate-busters \cite{Hasoc_2022_hatebuster} & 5 & 0.9592 & 0.9562 & 0.9568 \\
        4 & sakshi\_hasoc \cite{Hasoc_2022_sakshi} & 3 & 0.9464 & 0.9446 & 0.9450 \\
        5 & satlab \cite{Hasoc_2022_satlab} & 2 & 0.9372 & 0.9373 & 0.9372 \\
        6 & irlab@iitbhu \cite{Hasoc_2022_IRLab@IITBHU@IITBHU} & 5 & 0.9353 & 0.9355 & 0.9353 \\
        7 & ml\_ai\_iiitranchi \cite{Hasoc_2022_ml_ai_iiitranchi} & 2 & 0.9292 & 0.9203 & 0.921 \\
        8 & citk\_isi \cite{Hasoc_2022_CITK_ISI_ISI} & 1 & 0.9021 & 0.9022 & 0.9020 \\
        9 & gunjan & 3 & 0.8395 & 0.8071 & 0.7997 \\
        10 & the shivi hunters & 1 & 0.8053 & 0.7775 & 0.7746 \\
        \hline
    \end{tabular}
    }
    \caption{Results of Subtrack 3A. Ordered from best to least performing system.}
    \label{tab:3A_results_table}
\end{table}

\begin{table}[ht]
    \centering
    \scalebox{0.92}{
    \begin{tabular}{c|c|c|c|c|c}
        \hline
        \textbf{Rank} & \textbf{Name} & \textbf{Number of Runs} & \textbf{Precision} & \textbf{Recall} & \textbf{F1} \\
        \hline
        
        1 & hate-busters \cite{Hasoc_2022_hatebuster} & 5 & 0.9114 & 0.9342 & 0.9207 \\
        2 & satlab \cite{Hasoc_2022_satlab} & 2 & 0.9060 & 0.9288 & 0.9149 \\
        3 & ssncse\_nlp \cite{Hasoc_2022_ssncse_nlp} & 4 & 0.7587 & 0.7067 & 0.6958 \\
        4 & irlab@iitbhu \cite{Hasoc_2022_IRLab@IITBHU@IITBHU} & 1 & 0.4837 & 0.6107 & 0.5354 \\
        5 & ml\_ai\_iiitranchi \cite{Hasoc_2022_ml_ai_iiitranchi} & 2 & 0.4516 & 0.4371 & 0.4429 \\
        6 & citk\_isi \cite{Hasoc_2022_CITK_ISI_ISI} & 1 & 0.3405 & 0.2869 & 0.3073 \\
        7 & gunjan & 1 & 0.1111 & 0.2073 & 0.0897 \\

        \hline
    \end{tabular}
    }
    \caption{Results of Subtrack 3B. Ordered from best to least performing system.}
    \label{tab:3B_results_table}
\end{table}

\begin{table}[ht]
    \centering
    \scalebox{0.92}{
    \begin{tabular}{c|c|c|c|c|c}
        \hline
        \textbf{Rank} & \textbf{Name} & \textbf{Number of Runs} & \textbf{Precision} & \textbf{Recall} & \textbf{F1} \\
        \hline
        
        1 & satlab \cite{Hasoc_2022_satlab} & 2 & 0.9365 & 0.9893 & 0.9607 \\
        2 & ssncse\_nlp \cite{Hasoc_2022_ssncse_nlp} & 4 & 0.7963 & 0.8663 & 0.7929 \\
        3 & ml\_ai\_iiitranchi \cite{Hasoc_2022_ml_ai_iiitranchi} & 2 & 0.7927 & 0.7464 & 0.7423 \\
        4 & hate-busters \cite{Hasoc_2022_hatebuster} & 5 & 0.7999 & 0.7705 & 0.7255 \\
        5 & irlab@iitbhu \cite{Hasoc_2022_IRLab@IITBHU@IITBHU} & 1 & 0.2858 & 0.4213 & 0.2891 \\
        6 & citk\_isi \cite{Hasoc_2022_CITK_ISI_ISI} & 1 & 0.2323 & 0.1961 & 0.2064 \\

        \hline
    \end{tabular}
    }
    \caption{Results of Subtrack 3C. Ordered from best to least performing system.}
    \label{tab:3C_results_table}
\end{table}


Team ssncse\_nlp \cite{Hasoc_2022_ssncse_nlp} achieved the highest F1 of 0.974 for Subtrack 3A. Their RF model surpassed the performance of several transformer-based models, optimize\_prime \cite{Hasoc_2022_Optimize_Prime} and hate-busters \cite{Hasoc_2022_hatebuster}, that came 2nd and 3rd place respectively. Other traditional approaches also performed on par with deep learning approaches for this subtrack, with the logistic regressors submitted by satlab \cite{Hasoc_2022_satlab} and ml\_ai\_iiitranchi \cite{Hasoc_2022_ml_ai_iiitranchi} achieving similar performances to the transformer-based models provided by irlab@iitbhu \cite{Hasoc_2022_IRLab@IITBHU} and citk\_isi \cite{Hasoc_2022_CITK_ISI}. The size of the provided dataset is a possible explanation. Deep learning approaches were likely unable to infer useful feature representations that are able to differentiate between offensive and non-offensive instances, given that the provided dataset only contains 1067 offensive and 2034 non-offensive tweets. The submitted traditional approaches, on the other hand, were all engineered with features, such as token frequency or character-level n-grams, that have been proven to be useful for hate speech and offensive language identification \cite{mandl2019overview, mandl2020, hasoc2021mergeoverview}. Those deep learning approaches that did perform well, namely optimize\_prime \cite{Hasoc_2022_Optimize_Prime} and sakshi\_hasoc \cite{Hasoc_2022_sakshi}, also increased the amount of available training data by using augmented datasets. These augmented datasets included data from HASOC-2021 \cite{hasoc2021mergeoverview} or oversampled instances \cite{Hasoc_2022_sakshi} that likely increased their models' ability to infer distinguishing features.



Team hate-busters \cite{Hasoc_2022_hatebuster} came 1st place with an F1 of 0.921 for Subtrack 3B, closely followed by satlab \cite{Hasoc_2022_satlab} having achieved an F1 of 0.915. Team hate-busters was the only team that treated all three subtracks as one multi-class classification problem and that adopted an ensemble architecture. Their ensemble of L3Cude models in particular may be responsible for their high performance, as their final target output labels were generated by averaging those of multiple transformers trained on 10-folds.  This is further supported by the performance of the non-ensemble-based deep learning approaches submitted by irlab@iitbhu \cite{Hasoc_2022_IRLab@IITBHU} and citk\_isi \cite{Hasoc_2022_CITK_ISI}, with the latter team's submission also being pre-trained on L3Cude data. These teams scored drastically worst F1s of 0.5354 and 0.3073 respectively. However, the character-level n-gram features used by satlab appeared to have performed equally as well as hate-busters's ensemble. The use of such features as n-gram length, sentence position, and BM25 frequency, were likely able to identify character combinations or words, including pronouns, that are useful for the identification of targeted instances \cite{Hasoc_2022_satlab}. In addition, it appears that satlab did not remove meta data, such as “\textit{@USER}”, that would likewise have been beneficial; many of the other submissions removed such data during their pre-processing \cite{Hasoc_2022_Optimize_Prime, Hasoc_2022_IRLab@IITBHU}.




The 1st, 2nd, and 3rd leading teams of Subtrack 3C all used traditional approaches. The logistic regressor of satlab \cite{Hasoc_2022_satlab} came first place with F1 of 0.9607, whilst the RF and SVM of ssncse\_nlp \cite{Hasoc_2022_ssncse_nlp} and ml\_ai\_iiitranchi \cite{Hasoc_2022_ml_ai_iiitranchi}, came 2nd and 3rd place with F1s of 0.7929 and 0.7423 respectively. The deep learning approaches submitted to this subtrack all achieved inferior performances with the ensemble submitted by hate-busters \cite{Hasoc_2022_hatebuster} attaining the best F1 of 0.7255 and the non-ensemble-based models submitted by irlab@iitbhu \cite{Hasoc_2022_IRLab@IITBHU} and citk\_isi \cite{Hasoc_2022_CITK_ISI} getting noticeably worst F1s of 0.2891 and 0.2064 respectively. Traditional approaches were therefore superior for Subtrack 3C. A highly likely explanation for this is once again the size of the provided dataset. Subtrack 3C was a multi-class classification task, whereby models had to differentiate between three types of target for each instance: individual (IND), group (GRP), or other (OTH). However, only 503 instances with an individual target, 157 instances with a group target, and 80 instances with a target marked as other, were provided to the participating teams (Table \ref{tab_labels}). A recommended approach for dealing with such small sample sizes is to apply cross-lingual transfer learning for transformer-based models, such as training transformers on large datasets in Hindi and applying said models to Marathi, as shown by the winning system of HASOC-2021 \cite{nene-2021-transformer}. 

\section{Conclusion and Future Work} \label{conclusion}

This paper has described Subtrack 3 of HASOC-2022: Offensive Language Identification in Marathi. Subtrack 3 consisted of three additional subtracks for Marathi concerned with first distinguishing between hateful and non-hateful tweets, and then identifying whether a tweet had a target as well as its target type. In total, 10 teams submitted systems to this task and those that submitted system reports have been described in Section \ref{approaches}. Team performances have been listed and discussed in Section \ref{results}. In general, traditional approaches achieved the best performances, especially for Subtrack 3C. A RF (ssncse\_nlp) came 1st, 3rd, and 2nd place for subtracks 3A, 3B and 3C respectively \cite{Hasoc_2022_ssncse_nlp}, and a logistic regressor trained on n-gram features (satlab) came 2nd place for Subtrack 3B and 1st place for Subtrack 3C \cite{Hasoc_2022_satlab}. Deep learning approaches performed less well. However, those that that utilized an augmented dataset (optimize\_prime and sakshi\_hasoc) \cite{Hasoc_2022_Optimize_Prime, Hasoc_2022_sakshi}, or that had an ensemble architecture (hate-busters) \cite{Hasoc_2022_hatebuster} outperformed those that used a singular pre-trained deep learning model (irlab@iitbhu and citk\_isi) \cite{Hasoc_2022_IRLab@IITBHU, Hasoc_2022_CITK_ISI}. We believe this to be due to the small size of the provided dataset. 

Marathi continues to be an under-resourced language for many NLP-related tasks. Subtrack 3 aimed to encourage the development of ML models that were capable of performing well, regardless of having a limited amount of available data. The results of Subtrack 3 indicate that traditional approaches can still be effective when given a small training set. Nevertheless, ensemble architectures and cross-lingual transfer learning would likely surpass the performance of traditional approaches if said training sets were supplemented with additional instances in Marathi or Hindi. We suspect these to become common approaches within future iterations of this shared-task.

\begin{acknowledgments}
We would like to thank the teams referenced in this report for participating in the competition. We further thank the organizers of the other HASOC tracks for their support. 
\end{acknowledgments}

\bibliography{sample-ceur, kai_references}




\end{document}